\pdfoutput=1

\documentclass[11pt]{article}

\usepackage{ACL2023}

\usepackage{times}
\usepackage{latexsym}

\usepackage[T1]{fontenc}

\usepackage[utf8]{inputenc}

\usepackage{microtype}

\usepackage{inconsolata}

\usepackage{graphicx} %
\usepackage{amsmath}
\usepackage{multirow}
\usepackage{makecell}

\usepackage{pifont}%
\newcommand{\cmark}{\ding{51}}%
\newcommand{\xmark}{\ding{55}}%

\newcommand{\changeB}[1]{\textcolor{black}{#1}}

\title{Resolving Indirect Referring Expressions for Entity Selection}

\author{Mohammad Javad Hosseini \quad Filip Radlinski \quad Silvia Pareti \quad Annie Louis \\
        Google Research \\
        \texttt{\{javadh,filiprad,spareti,annielouis\}@google.com}
}

\begin{document}
\maketitle
\begin{abstract}
Recent advances in language modeling have enabled new conversational systems. In particular, it is often desirable for people to make choices among specified options when using such systems. 
We address this problem of reference resolution, when people use natural expressions to choose 
between the entities. For example, given the choice `Should we make a Simnel cake or a Pandan cake?'~a natural response from a dialog participant may be 
\emph{indirect}: `let's make the green one'. 
Such natural expressions have been little studied for reference resolution. We argue that 
robustly understanding such language has large
potential for improving naturalness in dialog, recommendation, and search systems.
We create {\tt AltEntities}\footnote{Our dataset can be found at \url{https://github.com/google-research-datasets/AltEntities}} (Alternative Entities), a new public dataset of 42K entity pairs and 
expressions (referring to one entity in the pair), and develop models for the 
disambiguation problem. Consisting of indirect referring expressions  
across three domains, our corpus enables for the first time the study of 
how language models can be adapted to this task.
We find they achieve $82\%$-$87\%$ accuracy in realistic settings, which while reasonable also invites further advances.

\end{abstract}

\section{Introduction}

Natural dialog often requires resolving referring expressions (REs), not only within and across texts, but also for grounding natural language 
expressions to specific entities or images. 
We focus on a specific conversational setting where a
speaker's utterance intends to disambiguate between known named
entities. While many aspects of RE resolution have been studied 
extensively, past work has focused on pragmatic reasoning \cite{dalereiter1995,frankgoodman2012}, 
influence of discourse  \cite{orita-etal-2015-discourse}, and multimodal (e.g., image) context \cite{zhang-vision2018}. 

In the specific case of dialog, when people make choices, the natural REs
are not always item names, spatial locations or attributes present in the question.
For instance when the choice is among items with similar names (perhaps disambiguating automatic speech recognition errors), or items with difficult to pronounce names, or where the user does not even recall which name is correct but instead recalls some higher level attribute, the user may choose an \emph{indirect} expression (Table \ref{tab:intro_examples}).
Most related to our work, \citet{celikyilmaz-etal-2014-resolving}
previously studied REs in response to a set of related items (e.g., Harry Potter movies) 
shown in a user interface. Their work both contains direct (using entity name), indirect, as well as
locational (entity's position on the screen) expressions.  
Predating recent advances in language models (LMs), their best model is a decision tree classifier consuming knowledge graph metadata.

\begin{table}[t!]
    \centering
    \begin{tabular}{|l|}
    \hline
    Did you mean a Simnel or {\bf Pandan} cake?\\ \hline
\emph{It looks surprisingly green in color} \\
\emph{Without any frosting or fruit}\\
\emph{It is made from some leaf}\\
\emph{Comes from Indonesia}\\
\emph{Isn’t the Easter one}\\ \hline
    \end{tabular}
    \caption{Responses to the question which intend to choose Pandan cake over the alternative.}
    \label{tab:intro_examples}
    \vspace{-4mm}
\end{table}

In this work, we created the {\tt AltEntities} corpus by a multi-step process, soliciting crowdworkers to provide diverse yet \emph{realistic} natural expressions for selecting entities in three domains: {\sc books}, {\sc recipes}, and {\sc music}. 
To obtain natural and casual dialogic language, we introduce a novel cartoon-based annotation approach (Figure \ref{fig:cartoon_completion_screen1}). 
{\tt AltEntities} consists of 6,247 alternative questions (presenting two entities) along with 42,529 REs.
In this context, REs are typically definite noun phrases with a pronominal head and a restrictive relative phrase or one of its reduced variants.

Our experiments are based on fine-tuned BERT \cite{devlin-etal-2019-bert} and T5 \cite{raffel2020exploring} LMs. We assess the representation of entity names as well as other sources of entity information. We find that
the results depend significantly on the \emph{type} of entity information provided to the models alongside the REs: If a LM only has access to the entity names but no other information, a case that might happen especially for long tail entities, accuracy is around $60\%$. On the other hand, if a LM is (unrealistically) given entity information that is identical to that shown to annotators producing the REs, accuracy is very high (up to $95\%$). However, if the model (more realistically) only has access to generic information that may or may not overlap with annotators' knowledge (Section \ref{sec:task}), accuracy of our models is only $82\%$-$87\%$, leaving significant room for methodological improvements.

\section{Related Work}
\label{sec:related_work}
Our work adds to recent efforts to allow users to speak more naturally to conversational systems. Here, we present the most related studies focusing on the properties of REs as well as their resolution.

{\bf Alternative Questions}. 
Our questions belong to the class of {\em alternative} questions (e.g. {\em `Are you staying or leaving?'}). Several studies have focused on the form and semantics of such questions, and differences from yes/no questions particularly  on the basis of prosody \cite{beck2006intervention,Biezma2012-BIERTA,pruitt2013interpretation}.

This paper focuses on the deep understanding of answers to such alternative questions when they are posed for selecting between two entities.

{\bf Speaker-Listener Cooperation.} 
The research in this space follow the Rational Speech Act Theory \cite{frankgoodman2012}, where the way speakers and listeners reason about each others' intentions and beliefs explains which attributes speakers pick to describe an entity, and how listeners disambiguate the entity. \newcite{vogel-etal-2013-implicatures,monroe2017} focus on the pragmatic reasoning involved during the conversation which helps in reaching a common understanding of the topic.
\newcite{wilkes1992coordinating} study how REs change as the conversation proceeds. In an experiment, they show that participants start from long and indefinite descriptions of images, but end up with short and definite references. \newcite{jordan2005learning} study the subproblem of content and attribute selection for generating object descriptions.

In our data collection, we assume a conversation between two humans in three dialog turns, where the first two turns prime the RE produced in the last turn (Section \ref{sec:data_collection}).

{\bf Common Ground}. In addition to the interlocutors' intentions, their prior or shared knowledge also plays an important role in how they understand each other's utterances. Sometimes the common knowledge arises from a shared situation, e.g., in navigation dialog \cite{engonopoulos2013predicting,misu-etal-2014-situated,fang2014collaborative} %
or the presence of a visual space \cite{Yu_2018_CVPR,bernardi2021linguistic}. In the latter, the common ground is given, i.e., it is assumed the image is what all participants in the interaction see in the same way. In many other situations, e.g., in a dialog between two friends about a movie or a book, the common ground is hidden and we can only make assumptions of what information participants share.

In this work, during data collection, we assume that annotators have access to rich common ground involving multiple modalities such as text, image, and video (Section \ref{sec:item_background}). During model training inference, we explore performance with varying levels of background information (Sectoin \ref{sec:models}).

{\bf Implicature Understanding}. This paper advances the broad area of understanding implicature in dialog. For example, a few recent papers developed datasets and models for indirect boolean responses (without saying `yes' or `no') \cite{pragst2018changing,louis2020id,takayama2021direct,damgaard-etal-2021-ill}. Interestingly, \newcite{ruis2022large} shows that LLMs cannot solve such implicatures in a zero-shot setting. 

{\bf RE resolution}.
There are few prior studies around the data and models for resolution tasks such as ours. \citet{stoyanchev2021action} built a method where references to items from prior context in a dialog are resolved by detecting state updates. Unlike our work, their REs focus on attributes (e.g., {\it Italian} in {\it the Italian restaurant}) discussed in prior dialog. \citet{celikyilmaz-etal-2014-resolving} collect REs to 
a target item among others shown on a screen (e.g., a set of Harry Potter movies). 
Their expressions contain both direct (reference to entity name) and
indirect references, where the latter comprise about 25\% of the data ($\approx 6$K REs). 
To aid the resolution of indirect ones, they include features
which capture the overlap between an expression and knowledge graph
attributes for each item.

Our work creates a large scale corpus ($42$K REs) exclusively for indirect REs, and explores how LMs encode the knowledge for disambiguation.

\section{Collecting Rich Referring Expressions}
\label{sec:data_collection}
To maximize generalizability, we collect data in three domains:  {\sc books}, {\sc recipes}, and {\sc music}. These were selected to cover a diverse variety of entity types with different kinds of available information --- e.g.~plot summaries for books, images for recipes, and lyrics and videos for songs. We performed careful and detailed annotations, and explain the annotation steps in this section.

\subsection{Cartoon-driven Annotation Setup}
\label{sec:crowd_task}

\begin{figure*}[ht]
    \centering
    \includegraphics[height=8cm,trim={0 4.8cm 0 0},clip]{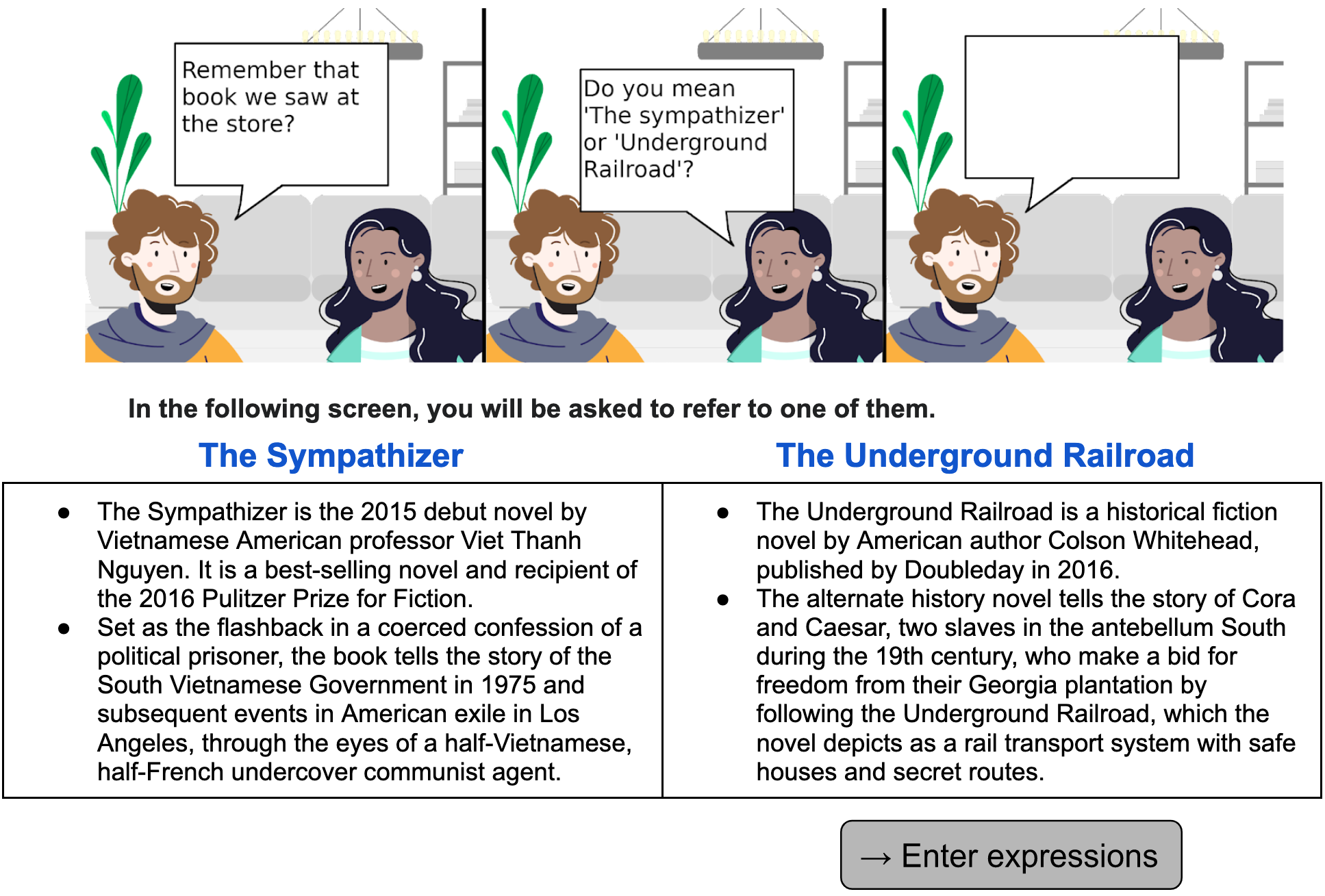}
    \caption{Annotators were shown a cartoon in which they were asked to complete the final step of a conversation.}
    \label{fig:cartoon_completion_screen1}
    \vspace{-4mm}
\end{figure*}

\begin{figure}[ht!]
    \centering
    \includegraphics[width=7cm]{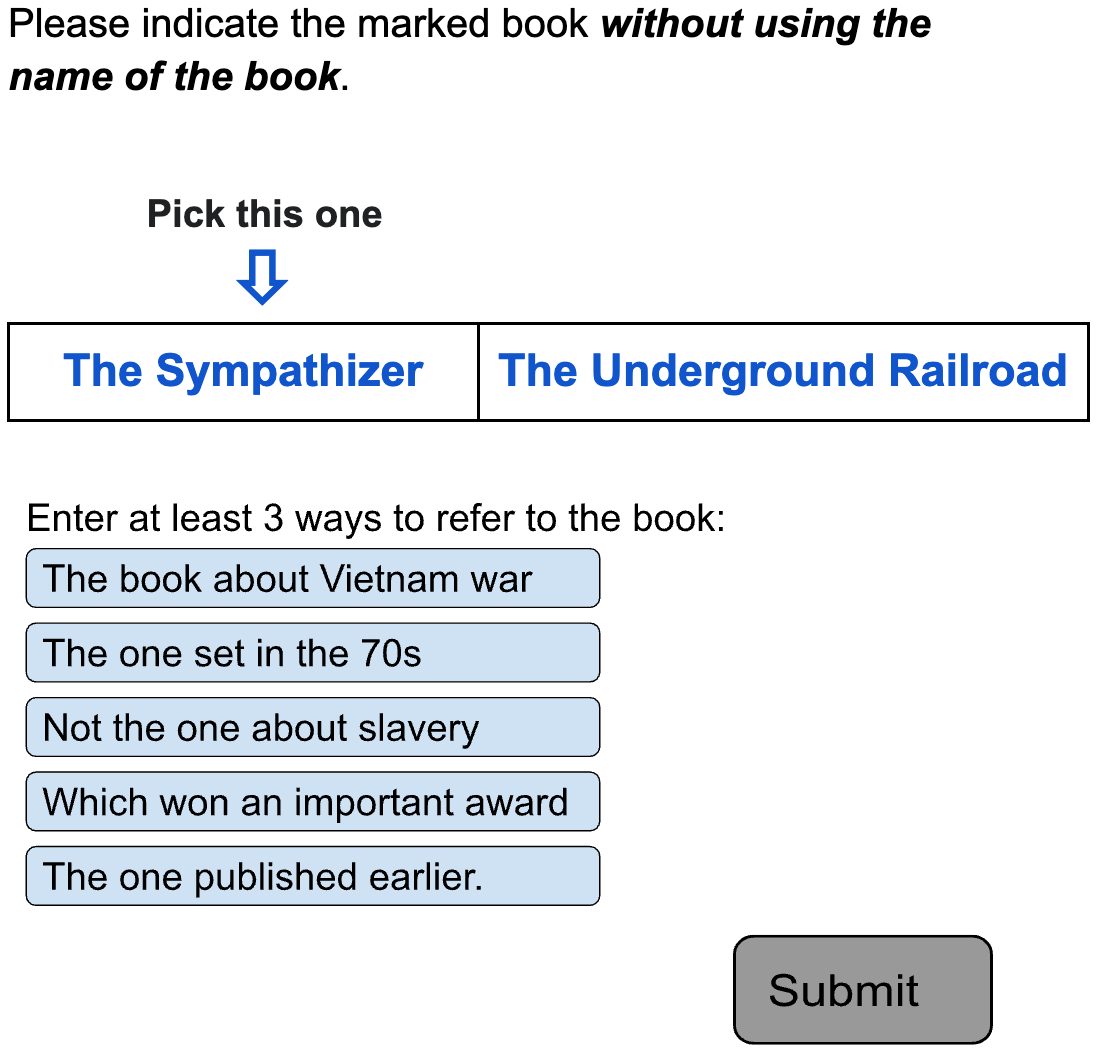}
    \vspace{-2mm}
    \caption{Annotation screen for entering expressions.}
    \label{fig:cartoon_completion_screen2}
    \vspace{-4mm}
\end{figure}

Previous work in question-answering and dialog typically asks annotators to complete text-based input boxes \cite{rajpurkar2016squad,choi2018quac,rajpurkar2018know,reddy2019coqa,eric2020multiwoz}. We employ a novel cartoon-bubble completion method, aiming to immerse annotators in the dialog setting to obtain more natural and informal REs. 
We start with a brief overview of the setup, and then explain the steps in detail.

Figure \ref{fig:cartoon_completion_screen1} shows the first (of our two) annotation screens. Annotators are shown a cartoon with two characters ({\em Bob} and {\em Alice})
in a fictional conversation, and asked (as Bob) to complete the last speech bubble.
This pictorial depiction, and the casting of
the dialog as a casual chat between friends encourage the annotators to produce friendly, short, and dialogic responses.
However, annotators are generally unlikely to know details about entities sampled from a collection. Therefore, we also provide background information on the entities (bottom of Figure \ref{fig:cartoon_completion_screen1}), corresponding to {\em common knowledge} that the two characters could share on the topic.

After annotators are shown this information, they proceed to a second screen (Figure~\ref{fig:cartoon_completion_screen2}). It indicates one of the entities (books in this example). They are asked to describe that entity (indirectly) with 3 to 5 responses: We found eliciting more entries encourages diversity and depth in the responses. Our data consists of the entity pairs, their descriptions, the target entity, and annotator expressions. 

From Figure \ref{fig:cartoon_completion_screen2}, note that once on the response screen, annotators cannot re-read descriptions. This encourages recall from memory. The reasoning behind this, and many other aspects of this design, are explained in the next sections.

\subsection{The Conversational Cartoon}
\label{sec:cartoon}
 
The cartoon has three cells as shown in Figure \ref{fig:cartoon_completion_screen1}. 
The first is a domain-specific utterance intended to set context. For example, {\em `Remember that book we saw at the store?'} sets up the dialog as one recalling a specific book. These utterances are from a set of five manually written expressions for each domain, with one selected at random for each conversation. Examples in the {\sc recipes} and {\sc music} domains are {\em `That recipe on today’s Masterchef was too good!'} and {\em `You sang that song really well yesterday.'} Appendix \ref{sec:opening-utterance} shows all these utterances.

The \emph{alternative} question is presented in the second cell. This question follows
a fixed template: {\it Do you mean `A' or `B'?} where `A' and `B' are the names of two \emph{related} entities. Our entities are sampled from Wikipedia page titles, with any disambiguation parentheses removed. When the names are identical, we retain the Wikipedia disambiguation: For instance, one such question is {\em Do you mean `The Gladiator (Turtledove novel)' or `The Gladiator (Scarrow novel)'?}.

The third cell is completed by the crowdworkers, assuming the role of {\em Bob} to enter text that refers to the target entity. They enter those expressions as shown in Figure \ref{fig:cartoon_completion_screen2}.
Further screenshots of our interface for all domains are provided in Appendix~\ref{sec:guidelines}.

\subsection{Entity Background}
\label{sec:item_background}

\begin{figure}[t]
    \centering
    \includegraphics[width=7.2cm]{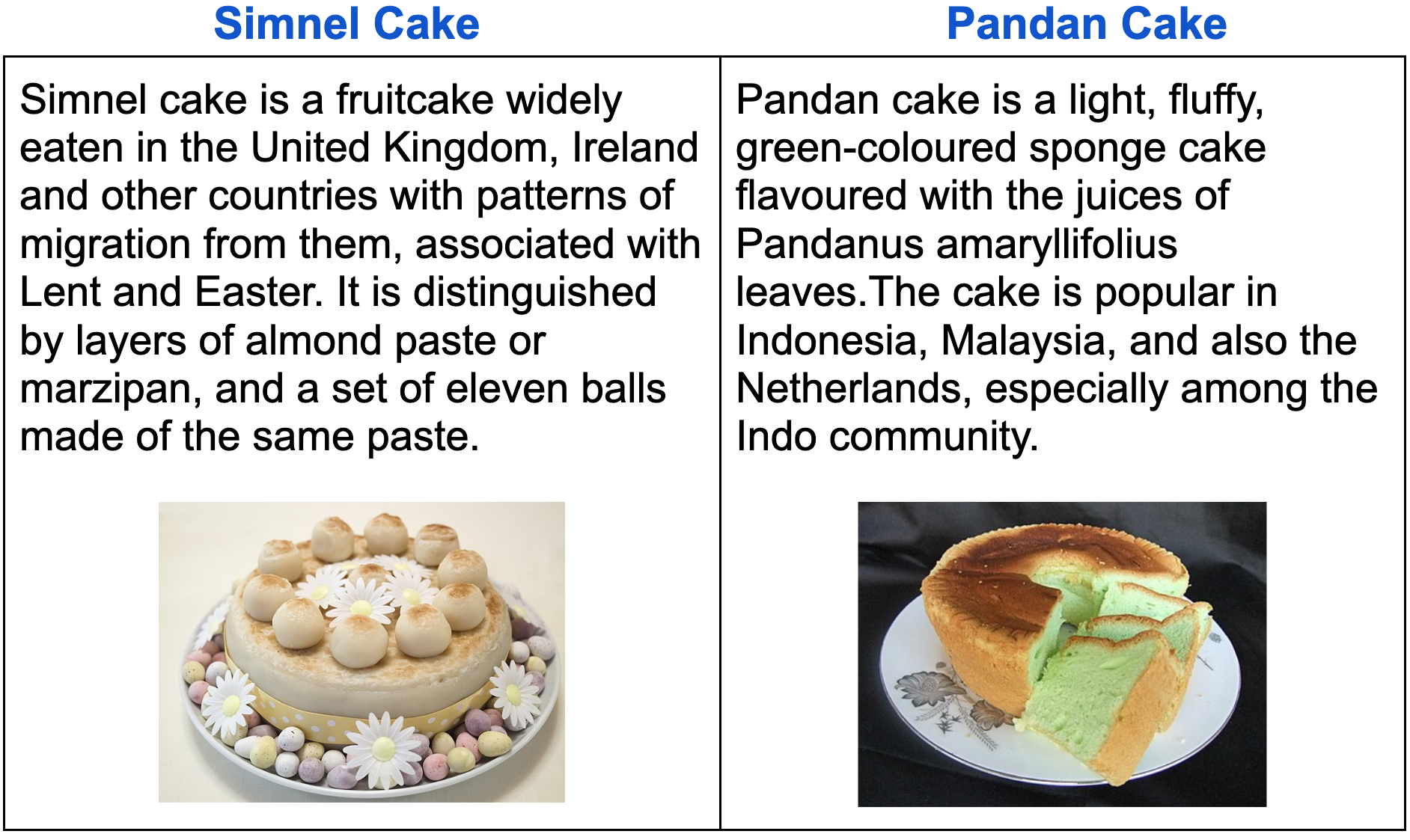}
    \caption{Background descriptions for two recipes.}
    \label{fig:recipe_description_example}
    \vspace{-4mm}
\end{figure}

In real dialogs, when people differentiate between options, they draw on partial knowledge about entities that they recall.
We aimed to foster a similar situation in our corpus, while doing so in a controlled manner without requiring domain-expert annotators. As such, when selected entities are shown to annotators, they are also presented with background information (bottom of Figure \ref{fig:cartoon_completion_screen1}). 
We draw the background also from Wikipedia, biasing towards sections relevant to each domain. 
For {\sc books}, these are the {\it main} (first) and {\it plot summary} sections. For {\sc recipes}, we used the {\it main}, {\it preparation}, and {\it ingredients} sections. 
For each entity, up to 750 characters of {\em one} of these sections are shown on the interface.
For {\sc recipes}, the food's image\footnote{We filtered out examples without any images.} is also always shown to help the annotators quickly realize what it looks like (Figure~\ref{fig:recipe_description_example}).%

For {\sc music}, however, we found Wikipedia text
to be less useful: Pages contain details and trivia (e.g., {\it 5th single on the album} or {\it sold 4 million copies}), which we judged 
unlikely to be included in natural background knowledge about a song. On the other hand, song lyrics and music are very relevant 
in this domain, but are not usually found in Wikipedia. Consequently, we presented a Google search link for the song in the 
background section, and asked the annotators to listen to at least some of each song, and read about them before writing
expressions. The search query contained the song's title and its artist, e.g.,~{\it Hello (by Adele)}. Since information
about the song comes from search, we also biased our candidates towards popular songs, which have more detailed results (Section \ref{sec:choices}).

\subsection{Generating Alternative Questions}
\label{sec:choices}

The alternative questions ({\it Do you mean `A' or `B'?}) are generated automatically: (i) Candidate entities are extracted from English Wikipedia for each domain (Section \ref{sec:candidates}), then (ii) we substitute `A' and `B' by sampling entity pairs (Section  \ref{sec:sampling_methods}).

\subsubsection{Selecting Candidate Entities}
\label{sec:candidates}

For each domain, we collect English Wikipedia articles by checking the presence of certain  Wikipedia templates (infoboxes\footnote{Infoboxes are fixed-format tables that consistently present articles in a given category (e.g., all books).}), and the presence of particular sections:
For {\sc recipes}, we additionally included articles with an {\it ingredients} section.

This set was then filtered to exclude very short articles, or those ambiguous between domains. 
For {\sc music}, we use article length (number of sections/subsections) as a proxy for 
popularity, and choose the top $\approx 1000$ articles. To remove any sensitive or offensive content, 
we also filter articles whose content matches a
list of sensitive words. Appendix \ref{sec:filter_articles} contains the details of the above filters. Table \ref{tab:items_count} shows the number of candidate entities.

\begin{table}[t!]
\centering
\small
\begin{tabular}{| c | c | c | c|} \hline
& {\bf \textsc{Books}} & {\bf \textsc{Recipes}} & {\bf \textsc{Music}}\\ \hline
{\bf Main} & 22,763& 2,822 & 1,032\\ \hline
{\bf Plot Summary} & 5,858 & - & - \\ \hline
{\bf Preparation} & - & 343 & - \\ \hline
{\bf Ingredients} & - & 147 & - \\ \hline
{\bf Total} & 28,621 & 3,312 & 1,032  \\ \hline
\end{tabular}
\normalsize
\caption{Number of extracted candidate items for each domain and background section.}
\label{tab:items_count}
\vspace{-4mm}
\end{table}

\subsubsection{Sampling Entity Pairs}
\label{sec:sampling_methods}

Much linguistic work on alternative questions has focused on the semantics and
pragmatics of these utterances \cite{Biezma2012-BIERTA}, but we also need to make decisions about which 
entity pairs could make for a challenging
disambiguation problem. Entity pairs sampled uniformly at random
are less likely to be interesting, since they may not share many properties, making disambiguation easier. In this work, we develop entity pair sampling techniques at different similarity levels, as a proxy for disambiguation difficulty.

{\bf Uniform sampling}. Entity pairs are sampled uniformly at random from the domain.

{\bf Same name}. These entities have the 
same name in Wikipedia followed by a disambiguation phrase within parentheses. An example is {\em Dawn (McLaughlin novel)} and {\em Dawn (Andrews novel)}. 

{\bf Similar title}. These entities have a similar
title in terms of character edit distance (distance $\leq3$), where the title could optionally consists of a disambiguation phrase within parentheses.

{\bf Similar description}. This method looks for deeper similarity within the text of
Wikipedia articles: We sample a first entity
uniformly, then select the second with the highest
similarity
using a Universal Sentence Encoder \cite{cer2018universal}. The
input to the encoder is the Wikipedia section shown as the background knowledge to annotators.

{\bf Similar infobox attributes}. Here we take
entities that share  important domain-specific properties, e.g., 
recipe origin, or the song genre. We match entities 
(except {\sc books}) using the `attributes' listed in the Wikipedia infobox: 
\{{\em type}\} and \{{\em type}, {\em country}\} for {\sc recipes}, and
\{{\em genre}\}, \{{\em artist}\}, and \{{\em genre}, {\em artist}\} for {\sc music}.

We applied the {\bf same name} method only to {\sc books}, and the {\bf similar title} method only to {\sc books} and {\sc recipes}. The other domains did not contain enough such examples. We applied the {\bf similar description} method to all domains. We applied the {\bf similar infobox attributes} method to {\sc recipes} and {\sc music}, but not the {\sc books} domain; however, some pairs with identical attributes were already covered by the other methods for {\sc books}. Table \ref{tab:entity_pairs_counts} shows the number of sampled entity pairs for each domain and sampling method.

\begin{table}[t!]
\centering
\small
\begin{tabular}{| l |  c |  c |   c |} \hline
& \textsc{Books} & \textsc{Recipes} & \textsc{Music}\\ \hline
Uniform & 649 & 813 & 700  \\ \hline
Same Name & 282 & - & - \\ \hline
Similar Title & 497 & 280 &	- \\ \hline
Similar Desc & 650 & 583 & 700 \\ \hline
Similar Attrs & - & 418 & 675 \\ \hline
All & 2,078 & 2,094 & 2,075  \\ \hline
\end{tabular}
\normalsize
\caption{Number of sampled entity pairs (questions) for each domain and sampling method.}
\label{tab:entity_pairs_counts}
\vspace{-3mm}
\end{table}

\subsection{Annotator Instructions and Pilot Runs}
\label{sec:instructions}

To maximize RE naturalness, we also
provided annotators different domain-specific examples. Figure~\ref{fig:cartoon_completion_screen2} shows those for 
the book {\it The sympathizer}. The REs are about topic ({\it about Vietnam war}), timeline ({\it set in the 70s}), and 
contrasts ({\it Not the one about slavery}, and {\it The one published earlier}).
They also emphasize use of general statements instead of 
overly specific and unrealistic ones,
e.g., {\it set in the 70s} instead of {\it 1975}. Table \ref{tab:books_do_list} shows a detailed note on desirable  expressions.

\begin{table}[t!]
\centering
\small
\begin{tabular}{| p{0.93\linewidth} |} \hline
\makecell{\bf Do} \\ \hline
\textcolor{teal}{\cmark~Keep it casual and conversational.}  \\ \hline
\textcolor{teal}{\cmark~Varied, interesting, and creative expressions.}  \\ \hline
\textcolor{teal}{\cmark~Use alternative words, e.g., {\it award} instead of {\it prize}. }  \\ \hline
\textcolor{teal}{\cmark~Vary the phrasing: {\it the book about}, {\it I meant the}, {\it was thinking of}, {\it the one about}, {\it I wasn’t referring to}, etc.}  \\ \hline
\makecell{\bf Don't} \\ \hline
\textcolor{red}{\xmark~Mention the book by name or position (e.g., the second one).}  \\ \hline
\textcolor{red}{\xmark~Use too detailed information that {\em Alice} may not recall (eg. {\it 1992} or {\it in the 90s} are better choices than {\it Sep 9 1992}).}  \\ \hline
\textcolor{red}{\xmark~Copy whole sentences from the description.}  \\ \hline
\end{tabular}
\normalsize
\caption{Actions annotators were encouraged (Do) or discouraged (Don't) to take for the {\sc books} domain.}
\label{tab:books_do_list}
\vspace{-4mm}
\end{table}

We performed pilot studies to understand how annotators responded to our instructions, and used these to refine the instructions. 
A first study (for {\sc books}) examined how annotators should use the background text, comparing designs where 
annotators could, or could not, go back-and-forth between the description screen (Figure~\ref{fig:cartoon_completion_screen1}), 
and the data collection screen (Figure \ref{fig:cartoon_completion_screen2}).
With back-and-forth possible, the responses contained excessive details, e.g., reiterating large portions of background text ({\it The book that was last of three juvenile novels that Wollheim wrote for Winston}).
With back-and-forth removed, annotators produced shorter REs ($7.99$ vs $9.61$ words), with fewer proper nouns and numbers per RE ($0.43$ vs $0.88$) as they are harder to remember.
They also used more contrastives, e.g., starting with {\em `not the'}
($21.8\%$ vs $2.2\%$) which involve drawing on information about both books. Thus, we adopted the 
memory recall setting.\footnote{Note that the {\sc music} entities are provided with search links which open in a new page, making back-and-forth possible, although it was discouraged in the guidelines.} After the first pilot study, we performed one pilot per domain for relatively small instruction refinements.

\section{The AltEntities Corpus}
\label{sec:altentities_corpus}

Our annotations were carried out using a pool of around $60$ in-house crowdworkers.\footnote{Paid contractors who work with our institution on such tasks.} They were all native English speakers recruited from U.S., U.K., Canada, and Australia so as to obtain a diverse set of perspectives.\footnote{The average number of questions per annotator is 217. The minimum number of annotations was 10, and the maximum was 2015 questions, followed by 610 questions. Around $80\%$ of annotators annotated around 100-600 questions each. We did not observe any obvious correlation between dataset artifacts and specific annotators.} Each question was shown to two workers to get multiple inputs per question. Around
$2$K entity pairs were annotated for each domain resulting in around $42$K expressions in total.
Table \ref{tab:dataset_analysis} shows the final corpus statistics, and Table \ref{tab:examples} shows example expressions for the three domains. We release the dataset under the CC-BY SA 3.0 License as per the Wikipedia License.

The REs for {\sc books} were on average a word longer than for other domains. They also contained more named entities per expression. Each domain contains some repeated REs (e.g., {\it the pop song}), that are often high-level responses, e.g., a song's genre. The {\sc books} domain contains the most unique responses.
The number of contrastives, estimated as REs starting with ``not the", are from $8\%$ in {\sc music} up to $20\%$ in {\sc books}.\footnote{This estimate gives a lower bound as there are other types of contrastives expressions
such as {\it the newer song}.}
For {\sc music} and {\sc recipes}, we manually checked $200$ random REs for references to modalities other than text. Around $10$\% multi-modal REs were present in the {\sc recipes} domain (mostly color), and $20$\% in the {\sc music} domain (mostly beat, speed, and mood).

\begin{table}[t!]
\centering
\small
\begin{tabular}{| l | c | c | c |} \hline
& {\bf \textsc{books}} & {\bf \textsc{recipes}} & {\bf \textsc{music}}\\ \hline
\# Questions & 2,078 & 2,094 & 2,075 \\ \hline
\# Expressions & 13,144 & 15,046 & 14,339 \\ \hline
Length (words) & 7.8  & 6.2 & 6.8 \\ \hline
\# Named Entities & 0.7  & 0.2 & 0.4 \\ \hline
Unique & 96\%  & 86\% & 76\% \\ \hline
Contrastives & 20\%  & 9\% & 8\% \\ \hline
Multi-modality & -  & 10\% & 20\% \\ \hline
Estimated Error rate & 4.5\%  & 6.7\%  & 6.8\% \\ \hline
\end{tabular}
\normalsize
\caption{The AltEntities corpus statistics}
\label{tab:dataset_analysis}
\vspace{-4mm}
\end{table}

We estimated the RE error rate by manually inspecting $40$ question samples (around $250$ to $300$ expressions) per domain. The error rate is between $4.5\%$ to $6.8\%$ for the three domains.
$78\%$ of these errors were due to the RE applying to both items, not just the target entity.
The remaining errors were mostly due to confusing the two entities.  We also note that the rate of exact string match between REs and Wikipedia text is $<1\%$.

The annotators were inspired by the provided stylistic cues in the instructions (e.g., starting with {\it the one} or {\it I meant the}), but followed our guidelines to vary their responses as well.
We observed that the content of REs (e.g., timeline, lyrics, singer or band information, instrument) included both the categories covered by the provided examples (e.g., timeline for books and songs) and novel categories (e.g., background information on books and songs such as {\it The one inspired by a Rolling Stones song}).

\begin{table}
\centering
\small
\begin{tabular}{| c |} \hline
{\bf \textsc{books}} \\ 
The one that is set in the 1880s \\
It's by a famous detective writer \\
The fictional one \\
not the one with the 12 year old boy \\
It's the book that has rock and politics in it \\ \hline 
{\bf \textsc{music}} \\
 The one without words \\
It is the song sung by an Australian. \\
It has synthesizer sounds in it  \\
Came out in mid of 2000. \\
Based on life experienced in Sheffield. \\ \hline 
{\bf \textsc{recipes}} \\
comes from Azerbaijan \\
The Japanese steamed cake \\
The ones eaten at Christmas \\
cornmeal is the main ingredient \\
Not the one with dried peaches. \\ \hline
\end{tabular}
\normalsize
\caption{Random REs from crowd annotators.}
\label{tab:examples}
\vspace{-4mm}
\end{table}
\section{Task and Models}
\label{sec:task}

Indirect reference resolution can be defined as follows: Given an alternative question with $K$ choices\footnote{In this paper, we only consider $K{=}2$.} $C=\{c_1,\ldots,c_K\}$, and a RE $r$, models should disambiguate the choice $c^* \in C$ intended by $r$.
We assume $r$ does not \emph{directly} mention $c^*$ by its name or position, but does uniquely \emph{refer} to $c^*$.

\subsection{Information Available to Models}

At a minimum, all models require the RE $r$ and the names of the choices $C=\{c_1,\ldots,c_K\}$. In addition, models may use textual descriptions $\{s_1,\ldots,s_K\}$ to aid disambiguation. We define choice text $s'_i$ ($1 \leq i \leq K$) as: (a) The entity name $c_i$,
or (b) the concatenation of $c_i$ and the textual description $s_i$, separated by a delimiter.\footnote{It is possible to use other modalities, e.g., recipe images or music videos; however we focus on text only.}
We consider the following four experimental setups.

{\bf \textsc{name}}: The entity name without further description of the entities.
We use this setting as a baseline.

For the remaining models, we add the following description to the name (truncated to $512$ tokens):

{\bf \textsc{InfoBox}}: The concatenation of all infobox key-value pairs (e.g., {\em `genre: pop'}).

{\bf \textsc{Unshown Background}}:  The \textsc{InfoBox} text, concatenated with all the Wikipedia sections of the entity, \emph{excluding} the section shown to the annotators as background. Since annotators were shown a search link and not a specific Wikipedia section for the {\sc music} domain, we do not remove any Wikipedia section for the {\sc music} entities. We note that the \textsc{Unshown Background} might have some overlap with the information shown to crowdworkers, but the text is not directly given to them. Hence, it is a fair setup to evaluate models in a practical system where the models might not have all the background information.

{\bf \textsc{Oracle}}: The same background text that was shown to the annotators (Section \ref{sec:item_background}). Note that this only exists for {\sc books} and {\sc recipes}, as for {\sc music}, annotators were only shown a search link.

\subsection{Models}
\label{sec:models}

We evaluated 5 different models. For each, we score match to each entity choices and select $c^*$ with the highest score value.

{\bf Universal Sentence Encoder}: We calculate the cosine similarity between the universal sentence encoder (USE; \citeauthor{cer2018universal}\citeyear{cer2018universal}) embeddings for the RE $r$ and each choice's text $s'_i$.

{\bf Entailment}: Using a textual entailment classifier, we classify whether 
a choice's text $s'_{i}$ entails the RE $r$. We use the confidence
of the  `entailment' label as the score. We use a BERT model trained on the MNLI  dataset \cite{mnli} as our classifier. For all models based on BERT, we use BERT large uncased.

{\bf BERT}. We turn our task into binary classification: We make one example per choice ($c_i$, $r$) with label 1 if $r$ refers to $c_i$; otherwise, label $0$. We finetune BERT with a binary classification layer (with two units) on top of its [CLS] token embeddings. The LM input is the sequence $[\text{CLS}]s'_i[\text{SEP}]r$.  
During inference, for each choice $c_i$, we compute the probability of label $1$ as its score.

{\bf BERT Joint}. In contrast to the above binary setup, we encode all the $K$ sequences $[\text{CLS}]s'_i[\text{SEP}]r$ with BERT. We apply a linear layer (with one unit) on top of the [CLS] token embeddings from each sequence. We normalize the scores using softmax. Finally, we minimize a categorical cross entropy loss given the $K$ scores. During inference, we directly use each choice's score.

{\bf T5}. We turn our task into binary classification, as with the BERT binary model. We fine-tune a T5 XL model (3B parameters) with input sequence ``expression: $r$ entity: $c_i$ description: $s_i$'' and output sequence $1$ or $0$. For the {\sc name} input type, the input sequence omits the ``description'' part.
\begin{table*}[t!]
\centering
\small
\begin{tabular}{| l | p{.69cm} | p{.69cm} |  p{.69cm} | p{.69cm} || p{.69cm} | p{.69cm} | p{.69cm} |  p{.69cm} ||  p{.69cm} |  p{.69cm} | p{.69cm} || p{.69cm} |}
\multicolumn{1}{c}{\textsc{ }}& \multicolumn{4}{c ||}{\bf \textsc{Books}}  & \multicolumn{4}{c ||}{\bf \textsc{Recipes}} & \multicolumn{3}{c||}{\bf \textsc{Music}} & \multicolumn{1}{c}{\textsc{ } } \\ \hline
& \textsc{Orac} & \textsc{Name} & \textsc{InBo} & \textsc{UnBa} & \textsc{Orac} & \textsc{Name} & \textsc{InBo} & \textsc{UnBa} & \textsc{Name} & \textsc{InBo} & \textsc{UnBa} & \textsc{Avg} \\ \hline
USE & 67.25 & 54.35 & 56.65 & 60.40 & 69.28 & 55.73 & 63.75 & 65.00 & \textbf{\textit{57.83}} & 61.05 & 60.08 & 61.03 \\ \hline
Entailment & 84.95 & 52.15 & 63.65 & 68.80 & 79.98 & 54.08 & 67.14 & 74.41 & 54.52 & 64.49 & 	71.84 & 66.91 \\ \hline
BERT & 93.30 & 50.55. & 74.35 & 79.80 & 87.87 & 53.32 & 77.84 & 81.01 & 53.93 & 61.60 & 73.13 & 71.52 \\ \hline
BERT Joint & \textbf{\textit{94.05}} & \bf{59.80} & 75.35 & \textbf{\textit{81.50}} & 88.94 & 54.12 & 75.21 & 80.87 & \textbf{\textit{56.59}} & 67.48 & 75.24 &	73.56 \\ \hline
T5 & {\bf 95.10} & 55.65 & {\bf 78.30} & {\bf 83.40} & {\bf 92.60} & {\bf 61.97} & {\bf 83.33} & {\bf 86.76} & {\bf 58.11} & {\bf 74.28} & {\bf 82.27} & {\bf 77.43} \\
\hline
\end{tabular}
\normalsize
\vspace{-2mm}
\caption{Indirect reference resolution results for different models on all domains and input types: {\sc Oracle} ({\sc Orac}), {\sc Name}, {\sc Infobox} ({
\sc InBo}), {\sc Unshown Background} ({\sc UnBa}). The best result of each column is boldfaced. When the difference between the best result and another result is not statistically significant (paired t-test with p-value < 0.05), the other result is made both bold and italic (only 4 cases).}
\label{tab:main_res}
\vspace{-3mm}
\end{table*}

\section{Experiments}
\label{sec:experiments}

We split the questions in the {\tt AltEntities} corpus in each domain into training (70\%), development (15\%), and test (15\%) sets. To avoid information leaking between the sets, we allow each {\it target} item to be in only one of the sets.
For the USE and entailment models, we do not tune any hyperparameters. For supervised models, we tune the learning rate, batch size, and number of epochs using a grid search on the development data ($96$ configurations for BERT and $24$ configurations for T5). We report the hyper-parameter details in Appendix \ref{sec:hyper-params-details}.

\subsection{Reference Resolution Accuracy}
\label{sec:main_res}

We compute the accuracy of
each (alternative question, RE) pair, i.e.~whether the correct choice is scored highest. As $K{=}2$ in our experiments, a random baseline has accuracy $50\%$.

We show the test set results in Table \ref{tab:main_res} for all domains and input types.\footnote{The development set results (Appendix \ref{sec:dev_set_results}) are slightly higher, but exhibit similar patterns.} For each model, we also show the average results of all input types. Among the models, USE performs worst ($61.03\%$), followed by the entailment model ($66.91\%$).
BERT Joint ($73.56\%$) is on average $1.61\%$ better than BERT ($71.52\%$), confirming that modeling the choices jointly is effective. T5 has the highest average results ($77.43\%$), as expected given that we experimented with T5 XL with 3B parameters compared to BERT large with 360M.

In the {\sc Oracle} setting for {\sc books} and {\sc recipes}, accuracy is understandably high (up to $95.10\%$ for {\sc books} and $92.60\%$ for {\sc recipes}). We note that these results are an over-estimate of the model capabilities.
On the other hand, in the {\sc name} setting, in most cases the results are slightly above $50\%$, with the best result being $61.97\%$ for the {\sc music} domain with the T5 model. Here the LMs rely on their memorized entity knowledge \cite{petroni2019language}, suggesting that BERT and T5 embeddings are not sufficient to resolve arbitrary entity references.

With the {\sc InfoBox} input, the T5 model accuracy is $78.30\%$, $83.33\%$ and $74.28\%$ for {\sc books}, {\sc recipes}, and {\sc music}, respectively. It increases to $83.40\%$, $86.76\%$, and $82.27\%$, respectively, with the {\sc Unshown Background} input where we add unstructured text data to the structured infobox data. This shows the text is helpful when resolving REs. In practical settings, models should work with relevant, but not necessary the same background knowledge as users because (1) it is not possible to have access to users' actual knowledge, and (2) models always have some limitation in the amount of text they can input. We thus rely on the {\sc Unshown Background} setting as a realistic setting for measuring the capabilities of the different models.

\begin{table}[t!]
\centering
\small
\begin{tabular}{c c | c |  c | c}
&& \multicolumn{3}{c}{\bf Test Domain} \\
&&\textsc{Books}& \textsc{Recipes} & \textsc{Music} \\ \hline
\multirow{3}{*}{\rotatebox[origin=c]{90}{\parbox{1.16cm}{\bf Training\\ Domain}}} & \textsc{Books} & 83.40 &	83.55 & 82.54 \\[1.3pt]
& \textsc{Recipes} & 81.60 & 86.76 & 82.96  \\[1.3pt] 
& \textsc{Music} & 82.05 & 84.80 & 82.27 \\[1.3pt] \hline
& \textsc{Mixed} & {\bf 83.90} & {\bf 87.47} & {\bf 83.28} \\[1.3pt] \hline
\end{tabular}
\normalsize
\caption{T5 results for the {\sc Unshown Background} setup, when trained on one domain and tested on another domain.}
\label{tab:cross_domain}
\vspace{-3mm}
\end{table}

\begin{table}[t!]
\centering
\small
\begin{tabular}{| l |  c |  c |   c |}
\multicolumn{1}{l|}{} & \textsc{Books} & \textsc{Recipes} & \textsc{Music}\\ \hline
Uniform & 90.30 & 92.54 & 88.58  \\ \hline
Same Name & 85.02 & - & - \\ \hline
Similar Title  & 83.86 & 86.29 &	- \\ \hline
Similar Desc & 74.70 & 82.24 & 80.39 \\ \hline
Similar Attrs & - & 81.55 & 77.12 \\ \hline
All & 83.40 & 86.76 & 82.27  \\ \hline
\end{tabular}
\normalsize
\caption{T5 results with different sampling methods for each domain with {\sc Unshown Background} input.}
\label{tab:sampling_mehtods_t5}
\vspace{-3mm}
\end{table}

\begin{table*}[ht!]
\centering
\small
\begin{tabular}{|m{0.22\linewidth} | m{0.22\linewidth}| m{.23\linewidth} | m{0.21\linewidth} |} \hline
{\bf Error Type} & {\bf Target Item} & {\bf Non-Target Item} & {\bf Annotator Utterance}\\ \hline
\multirow{2}{*}{\makecell[l]{No Textual Overlap \\ $47\%$(B) $27\%$(R) $42\%$(M)}} & {\bf Best Song Ever} is a song recorded by English-Irish... & {\bf These Days} is a song by British pop group... & It has to do something with dancing all night. \\\cline{2-4}
& {{\bf Boerewors}..., a type of sausage which originated in South Africa.}  & {\bf White pudding} is a meat dish popular in Ireland, Northern Ireland...   & It can be stewed. \\ \hline
\multirow{2}{*}{\makecell[tl]{Poor reasoning \\ $25\%$(B) $18\%$(R) $13\%$(M)}} & {\bf Clams casino} is a clam "on the halfshell" dish... & {\bf Buddha's delight} ... is a vegetarian dish... &  The one with seafood in sauce.\\\cline{2-4}
& {\bf Dark Age}... release\_date: July 30, 2019... &  {\bf Iron Gold}... release\_date: January 16, 2018... &  It is the most recent one.\\ \hline
\multirow{2}{*}{\makecell[tl]{Multi-modality \\ $0\%$(B) $25\%$(R) $22\%$(M)}}  & {\bf It's Not Over} is the debut single by American rock... & {\bf Love Child} is a 1968 song released by the Motown... & Has a marriage proposal in the music video \\ \cline{2-4}
& {\bf Pandoro} appeared in remote times, the product of... & {\bf Pandebono}... It is said that an Italian baker who lived...  & Brownish-yellow in its colour.\\\hline
\multirow{2}{*}{\makecell[tl]{Wrong Annotation \\ $28\%$(B) $30\%$(R) $23\%$(M)}} & {\bf My Story (Gillard book)} is a political memoir of Julia Gillard... & {\bf My Story (Das book)} is an autobiographical book written by Indian author... & I mean the book that is technically an auto-biography. \\\cline{2-4}
& {\bf Tight Connection to My Heart} (by Bob Dylan)... & {\bf Like a Rolling Stone} (by Bob Dylan)... & this song is by an American singer.\\ \hline
\end{tabular}
\normalsize
\vspace{-2mm}
\caption{Error analysis results. Under each error type, we report the percentage of examples from the {\sc books} (B), {\sc recipes} (R), and {\sc music} (M) domains. We also show two example for each error type.}
\label{tab:error_analysis}
\end{table*}

\subsection{Cross-Domain Experiments}
\label{sec:res_cross_domain}

Reference resolution is a semantic task, and ideally models 
would learn general task aspects rather than domain details. We
test generalization by finetuning our models on one domain and 
testing on another. We used the {\sc Unshown Background} setting for these experiments as the most realistic.

Table \ref{tab:cross_domain} shows the T5 model results.\footnote{We observe similar results with BERT Joint and BERT models, which are not shown due to space limitations.}
We do not observe much difference when models are tested out of domain, supporting the hypothesis that our models are indeed generalizable. This observation is rather important since our models could be used
without separate training for new choice domains.

We also create a {\em mixed} training (and development) set that combines the data of the three domains. The mixed training set gives better results on average, taking advantage of larger training set and cues from all the domains. However, since the dataset in each domain is relatively large, the mixed training does not increase the results substantially.

\subsection{Results and Entity Similarity}
\label{sec:res_sampling_methods}

Section \ref{sec:candidates} explained how we selected 
entity pairs to have different levels of similarity.
We now examine how this affects performance. 
Table \ref{tab:sampling_mehtods_t5} shows the results for  
the T5 model with the {\sc Unshown Background} input. We compute accuracy per 
test example subset, where each originated from
a specific similarity sampling method.

As expected, when the two entities are randomly selected,
disambiguation is easiest since they have little in common. The task becomes harder as entities become more similar, with entities with similar infobox attributes having the lowest performance.

\subsection{Error Analysis}
\label{sec:error_analysis}
We analyzed the errors from the T5 model in the {\sc Unshown Background} setting, to understand if there are systematic errors which could be improved upon in the future. We manually analyzed $40$ incorrectly predicted development set examples per domain. We show four different error types and their percentages per domain in Table \ref{tab:error_analysis}.

In most cases, there is no textual overlap between the RE and the background. This is because either the relevant text is removed (by design) since it is shown to the raters, or the Wikipedia text does not contain the information at all (e.g., music lyrics). Future research could evaluate how to adapt LMs to improve their entity knowledge to reason beyond the input textual evidence. In addition, retrieval augmented LMs could be applied to retrieve relevant information before performing the prediction \cite{borgeaud2022improving,shi2023replug}.

In other cases, the model suffers from poor reasoning, e.g., that clam is seafood, or a vegetarian dish does not contain seafood. In addition, the model often misclassifies examples when entity attributes are compared (e.g., {{\it the newer one}}).
Multi-modality covers around $25\%$ of the errors in the {\sc recipes} and {\sc music} domains, e.g., annotators referenced visual aspects from music videos or recipes (e.g., {\it looks like shells}), or an acoustic aspect from a song (e.g., {\it with the piano intro} or {\it more upbeat}).

The remaining errors are because of wrong annotations, usually with the REs appling to both items.
This wrong annotation rate ($23\%$-$30\%$) is much higher than the error rate in the whole dataset (less than $7\%$ as discussed in Section \ref{sec:altentities_corpus}) since the model has learned the task to a good extent.

We also analyzed correctly classified examples (for the {\sc music} domain) to understand what types of REs are classified correctly. The results are shown in Appendix \ref{sec:correct_analysis}.

\section{Conclusion}
We have revisited RE resolution with a new focus on indirect expressions, introducing {\tt AltEntities}, a new large dataset for this task -- covering {\sc books}, {\sc recipes}, and {\sc music} examples. The dataset was collected using a novel cartoon completion approach to encourage conversational and causal expressions while avoiding name or position expressions. %
The experimental results show that in a realistic setting, LMs adapted for this task achieve $82\%$-$87\%$ accuracy. While an improvement on existing approaches, this also encourages further research on this important problem. Moreover, we showed that the models' performance does not drop when trained and tested on different domains, suggesting that models can learn the semantic task well and generalize to new domains. 

It is notable that in practice, many entities do not have textual descriptions or rich meta-data. Future research could study resolving REs with minimal information, e.g., when we only have access to their names or limited meta-data.
Future research could also use multi-modal input for training and inference. Further, to handle more complex REs such as {\it the newer one}, or {\it the happy song}, one could decompose a RE into simpler expressions and then perform the comparison. Similar data collection methodologies could be applied to collect a dataset with more number of choices and also cases where neither or multiple choices match the RE.

\section{Limitations}
\label{sec:limitations}
As with any natural language understanding task, there are practical limitations and related ethical aspects that must be considered before deploying a  system. In particular, our corpus and modeling approach assume that the user-provided REs \emph{always} refer to one of the two options. If this is not the case, or if the RE is particularly contrived, undesirable or unexpected behavior may occur: For any expression, including for instance one made with arbitrary derisive language, the model would attempt to resolve this to one of the alternative entities. One approach system designers may consider could be to pre-classify any user-provided REs to avoid interpreting those that are off topic or phrased in a negative manner.

A second consideration is that of corpus representativeness.
In our case, as this is a first corpus for this task, we have limited
ourselves to English Wikipedia, native English speaking annotators, and particular item sampling strategies for practical reasons.
However, if used for training a deployed system, the examples present may bias any model to understand specific types of references but not others. Similarly, the items in our corpus are sufficiently popular to have a relatively long Wikipedia entry, whereas items not present in Wikipedia, or with only minimal information, may exhibit different characteristics.

\section{Ethics Statement}
The data collection protocol was reviewed by an ethics panel to remove potential ethical concerns. A few ethical concerns were mentioned by the panel which were then judged to be handled well. These included ensuring that the entities, texts and REs were free from biased and sensitive language. We address this by filtering using a list of sensitive words (see Section \ref{sec:candidates} and Table \ref{tab:extraction_filters}). The panel also recommended a diverse representation of entities and domains. Thus our data comes from diverse domains and the entities are sampled from a large set of Wikipedia articles.

Still, we note that the limitations mentioned in Section \ref{sec:limitations} need to be considered and addressed carefully when using our dataset or models for evaluation or training of a deployed system. In addition, a biased corpus may lead to an evaluation that is unaware of RE language forms used in other cultures and languages, or that refer to other types of items. We expect this consideration to be important in practical settings.

\bibliography{anthology,indirect_reference}
\bibliographystyle{acl_natbib}

\clearpage
\pagebreak

\appendix
\appendix

\section{Opening Utterances}
\label{sec:opening-utterance}

The first annotation screen (Figure \ref{fig:cartoon_completion_screen1}) starts with a manually written opening utterance. Table \ref{tab:domain_questions} shows all these utterances for the three domains..

\section{Annotation Guidelines}
\label{sec:guidelines}
In this section, we provide the domain-specific guidelines that were shown to the annotators prior to the start of their annotation. The guidelines for each domain includes three {\it instruction} screens. The second and third instruction screens are then repeated for each alternative question as their first and second {\it annotation} screens, respectively (the two screen discussed in Section \ref{sec:altentities_corpus}).

\begin{figure*}
    \centering
    \frame{\includegraphics[width=15.5cm]{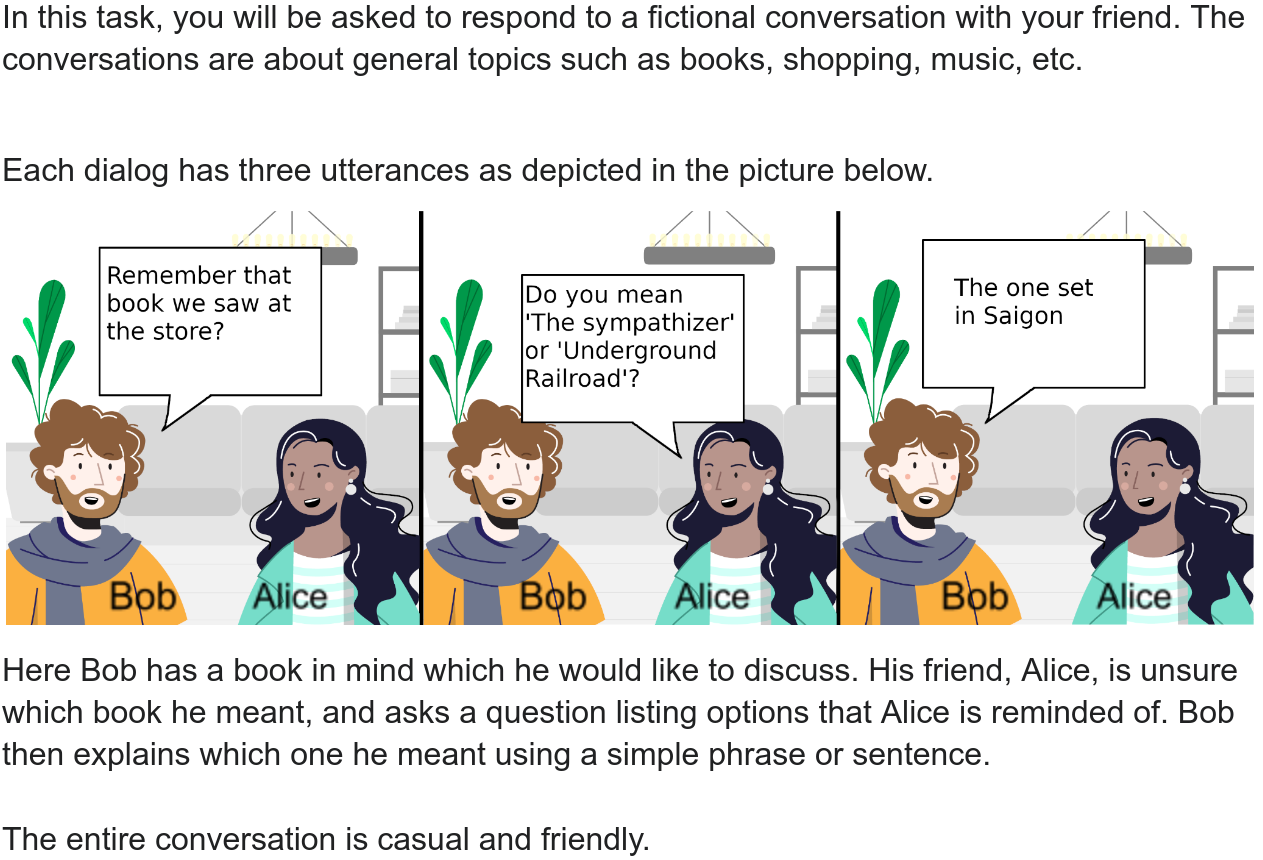}}
    \caption{The first instruction screen shown for the {\sc books} domain. It summarizes the task based on a cartoon completion setup.}
    \label{fig:books_first_screen}
\end{figure*}

In the first instruction screen, a summary of the task based on a cartoon completion setup is shown to the annotators. Figure \ref{fig:books_first_screen} shows the first instruction screen for the {\sc books} domain. We do not show the first instruction screen for the other two domains as they are very similar to the {\sc books} domain except that the text is slightly different to reflect the domain, and that the examples are from those domains.

\begin{figure*}
    \centering
    \frame{\includegraphics[width=15.5cm]{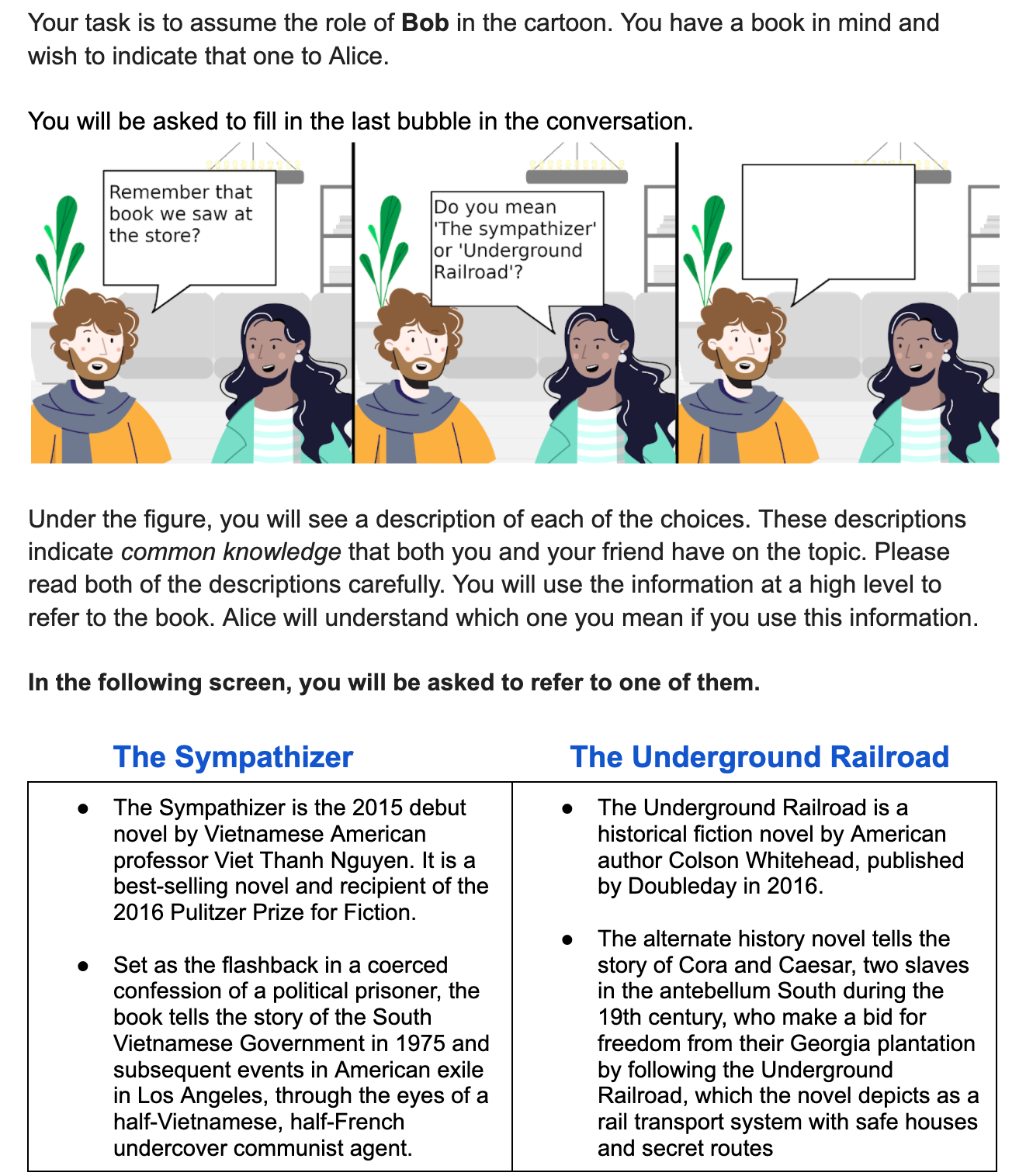}}
    \caption{The second instruction screen shown for the {\sc books} domain. It provides further information about the task and describes where the annotators should acquire the knowledge to perform the annotations.}
    \label{fig:books_second_screen}
\end{figure*}

\begin{figure*}
    \centering
    \frame{\includegraphics[width=15.5cm]{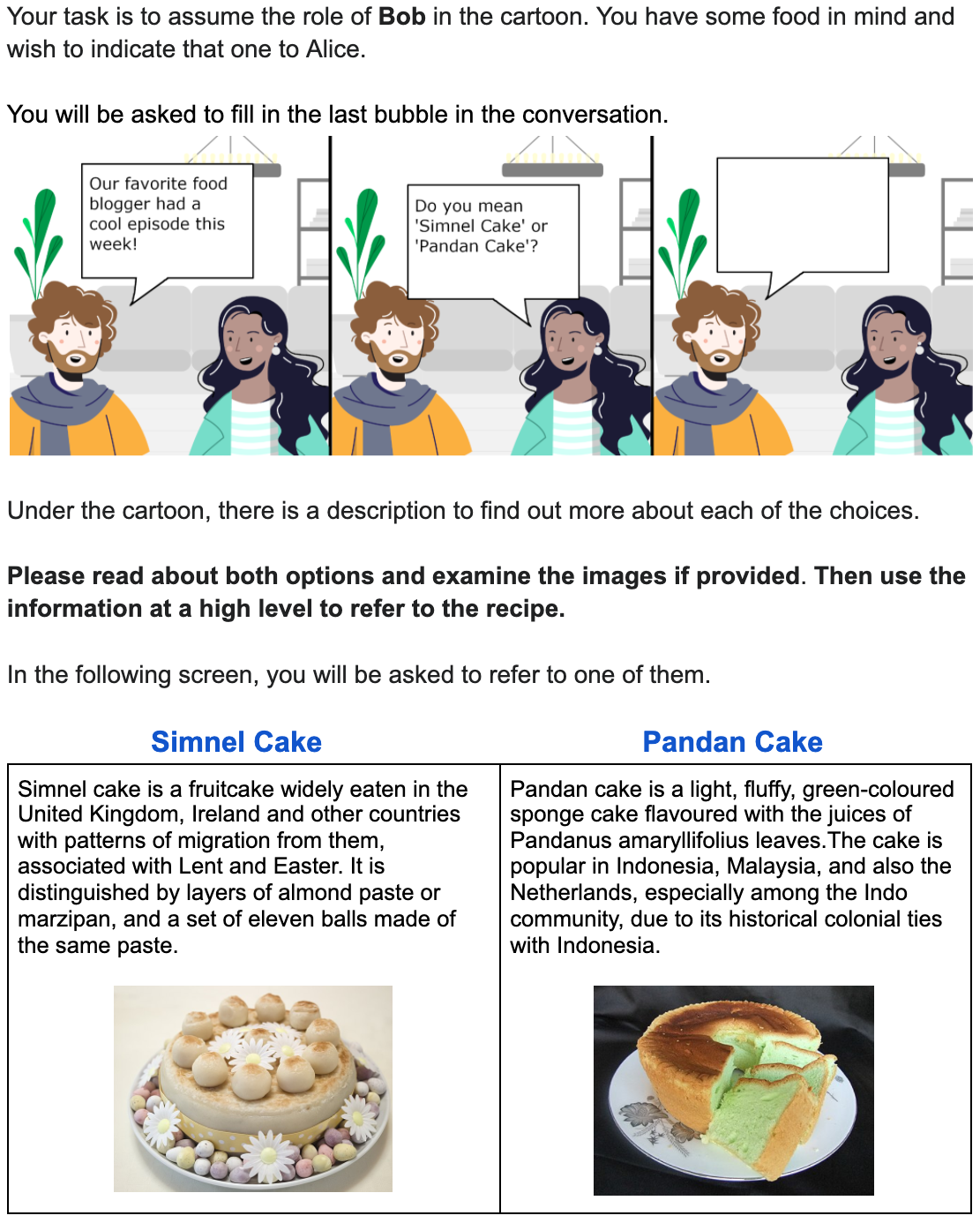}}
    \caption{The second instruction screen shown for the {\sc recipes} domain. It provides further information about the task and describes where the annotators should acquire the knowledge to perform the annotations.}
    \label{fig:recipes_second_screen}
\end{figure*}

\begin{figure*}
    \centering
    \frame{\includegraphics[width=15.5cm]{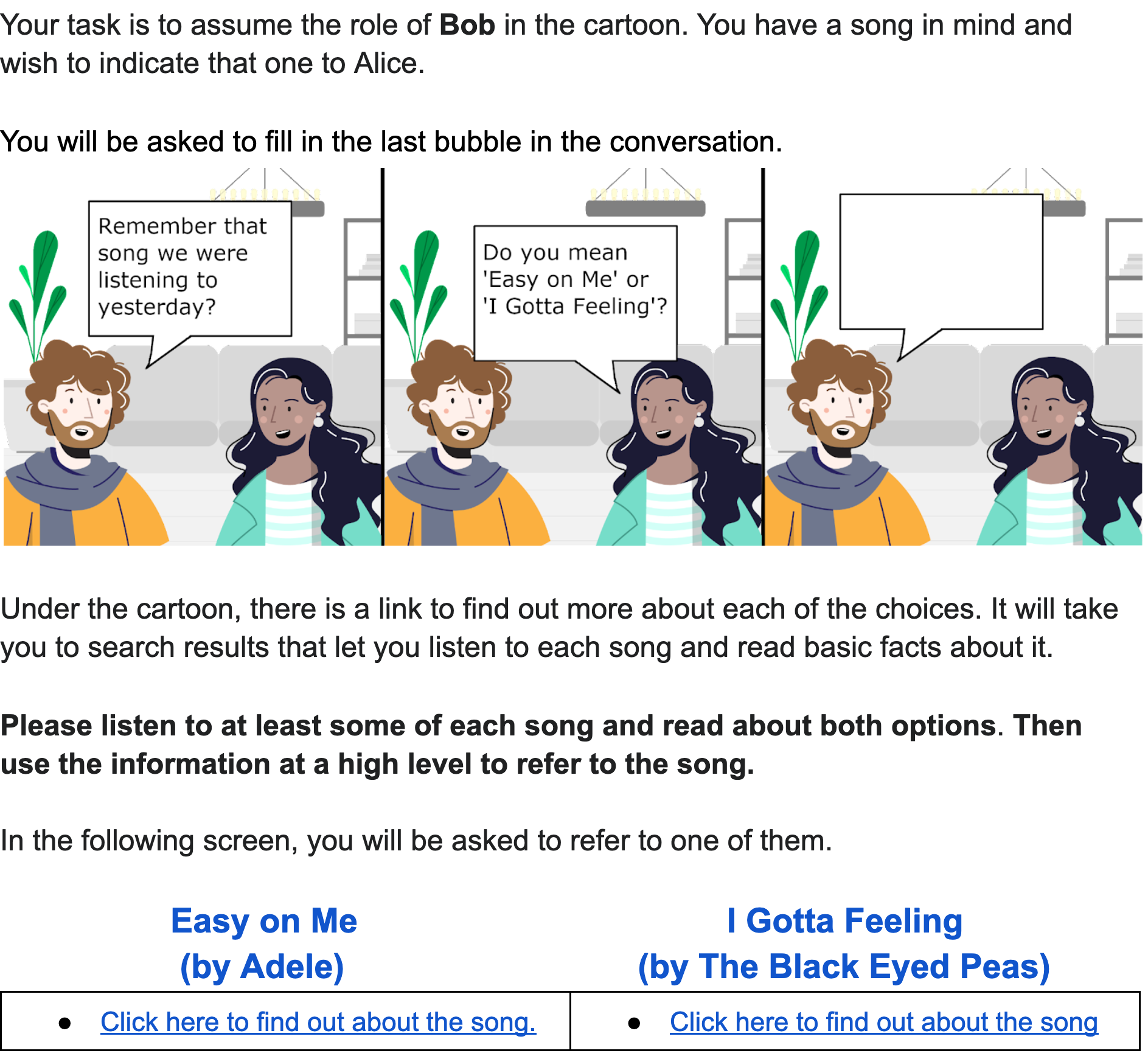}}
    \caption{The second instruction screen shown for the {\sc music} domain. It provides further information about the task and describes where the annotators should acquire the knowledge to perform the annotations.}
    \label{fig:music_second_screen}
\end{figure*}

The second instruction screen provides further information about the task and describes where the annotators should acquire the knowledge to perform the annotations. Figures \ref{fig:books_second_screen}, and \ref{fig:recipes_second_screen}, and \ref{fig:music_second_screen} show the second instruction screens for the {\sc books}, {\sc recipes}, and {\sc music} domains, respectively.

\begin{figure*}
    \centering
    \frame{\includegraphics[width=15.5cm]{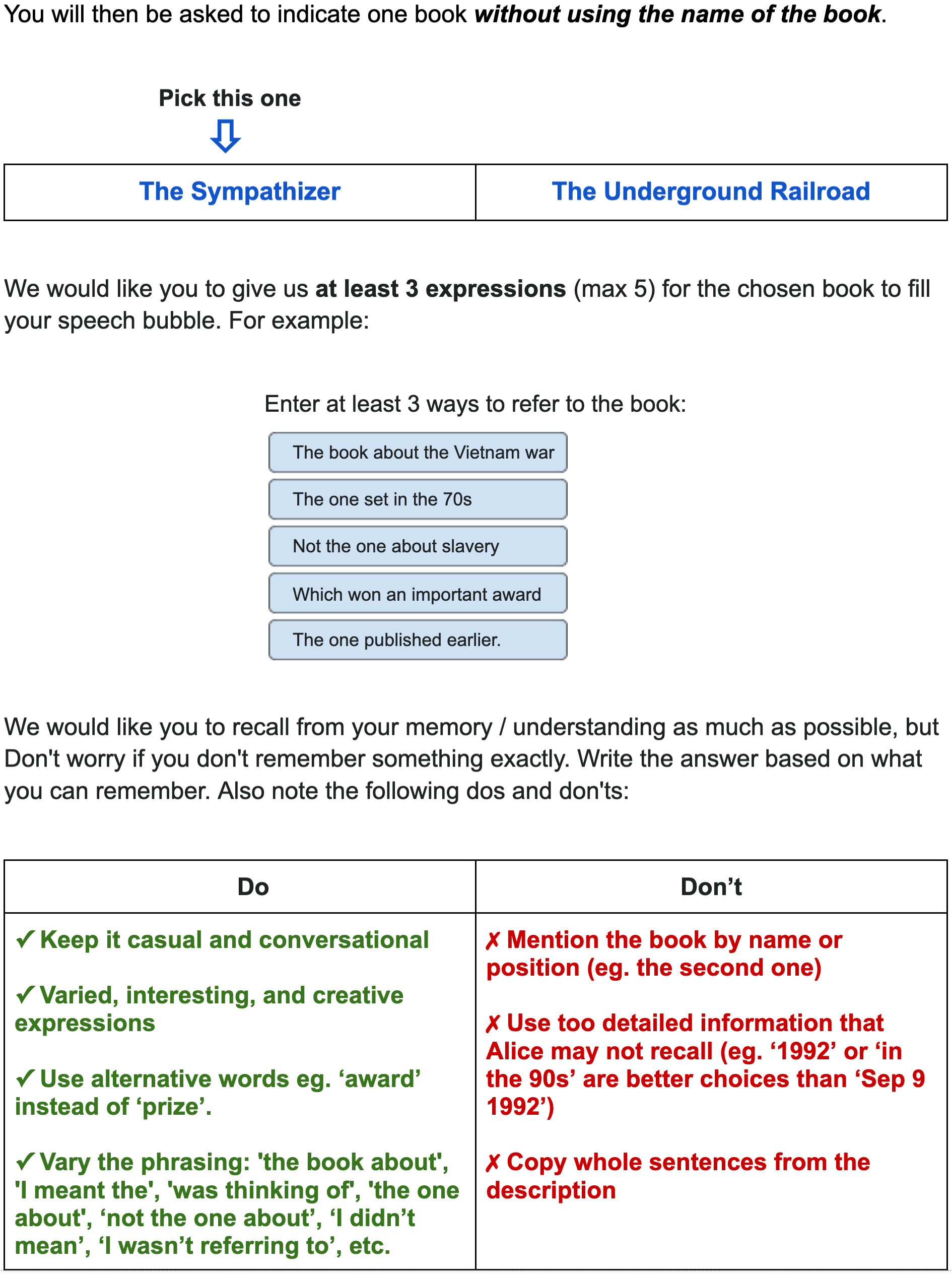}}
    \caption{The third instruction screen shown for the {\sc books} domain. It shows which item should be referred to, and lists five examples of appropriate REs. It also lists a number of actions that the annotators should or should not do.}
    \label{fig:books_third_screen}
\end{figure*}

\begin{figure*}
    \centering
    \frame{\includegraphics[width=15.5cm]{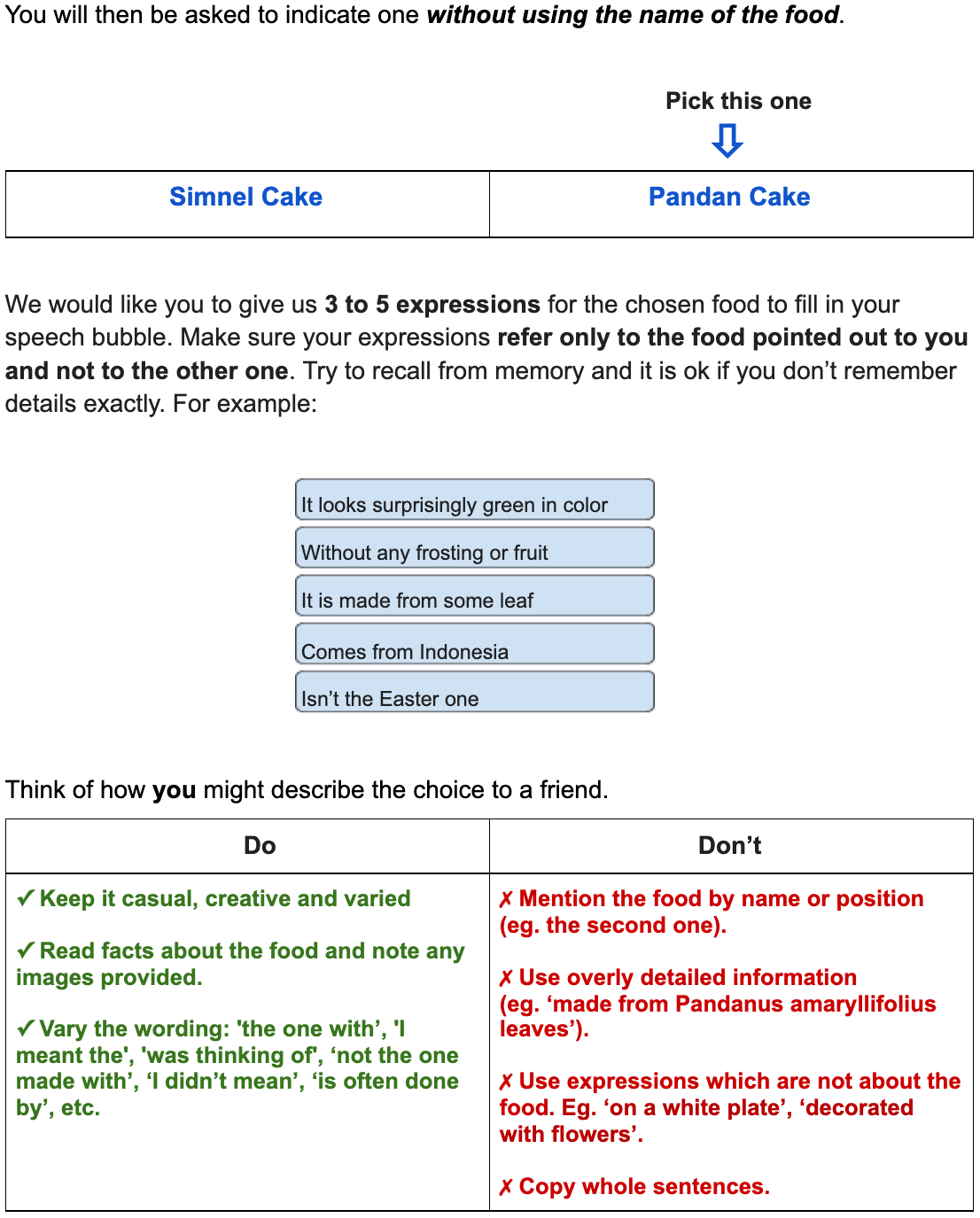}}
    \caption{The third instruction screen shown for the {\sc recipes} domain. It shows which item should be referred to, and lists five examples of appropriate REs. It also lists a number of actions that the annotators should or should not do.}
    \label{fig:recipes_third_screen}
\end{figure*}

\begin{figure*}
    \centering
    \frame{\includegraphics[width=15.5cm]{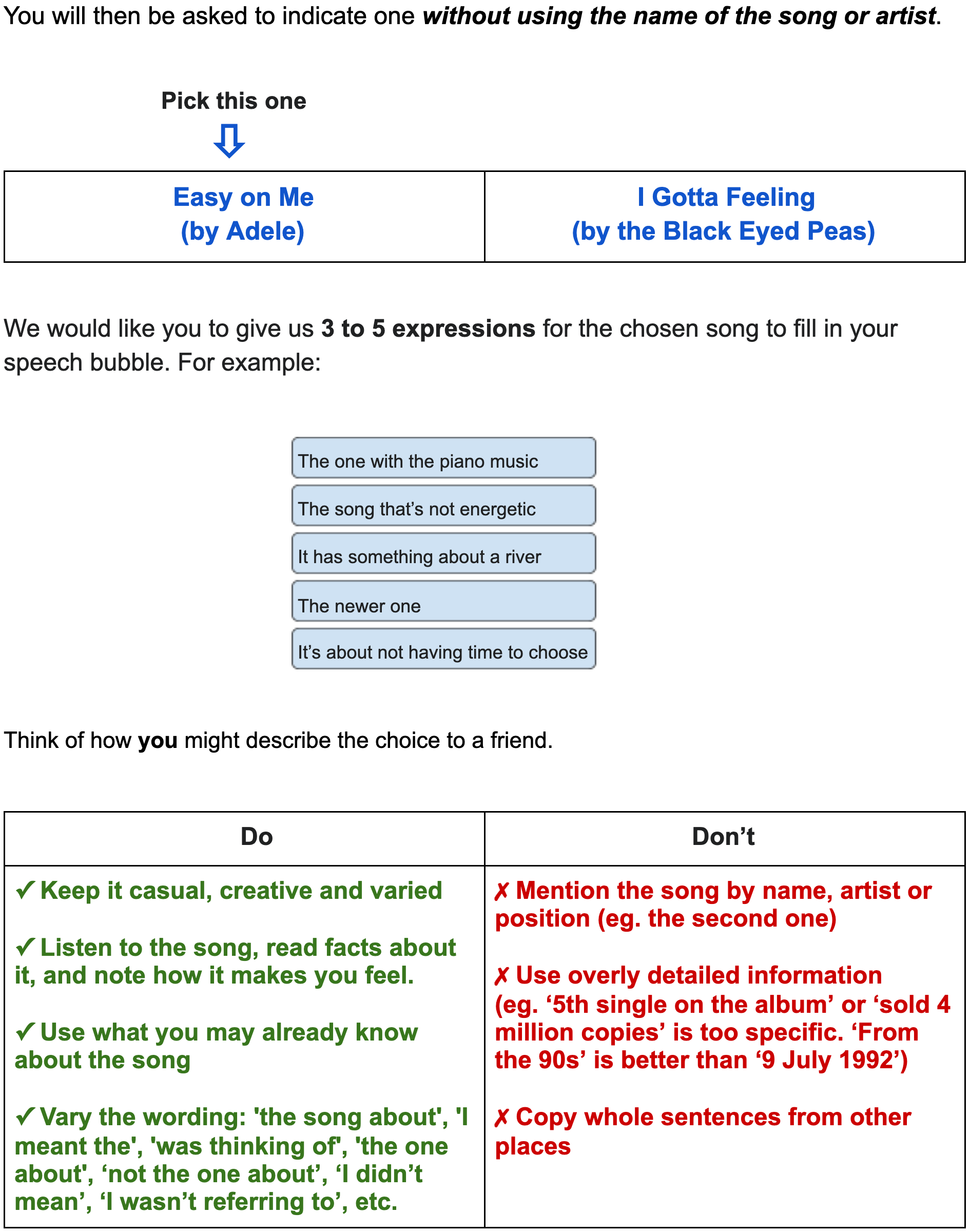}}
    \caption{The third instruction screen shown for the {\sc music} domain. It shows which item should be referred to, and lists five examples of appropriate REs. It also lists a number of actions that the annotators should or should not do.}
    \label{fig:music_third_screen}
\end{figure*}

The third instruction screen shows which item should be referred to, and lists five examples of appropriate REs. The REs cover different aspects of the items to encourage the annotators to cover a variety of the item aspects. It also lists a number of actions that the annotators should or should not do. Figures \ref{fig:books_third_screen}, \ref{fig:recipes_third_screen}, and \ref{fig:music_third_screen} show the third instruction screen for the {\sc books}, {\sc recipes}, and {\sc music} domains, respectively.

\section{Filtering Wikipedia Articles}
\label{sec:filter_articles}
Table \ref{tab:extraction_filters} shows a number of filters we applied to narrow down the extracted articles.

\section{Hyper-parameters Details and Computing Infrastructure}
\label{sec:hyper-params-details}

We tune the hyper-parameters using a grid search based on the accuracy of the indirect reference resolution task on the development set of each domain. For BERT and BERT multiple choice models, we select the base learning rate from $\{1e{-}4,5e{-}5,3e{-}5,1e{-}5,5e{-}6,3e{-}6,1e{-}6,\allowbreak 5e{-}7\}$, the training batch size from $\{16,32,64\}$, and the number of epochs from $\{1,3,5,10\}$. For T5, we select the base learning rate from $\{5e{-}7,1e{-}7,3e{-}6,5e{-}6,1e{-}5,3e{-}5,5e{-}5,\allowbreak 1e{-}4\}$ and the training batch size from $\{16,32,64\}$. We train the T5 models for 50K steps (batches).

Table \ref{tab:hyper-params} shows the selected hyper-parameters for each model, domain, and input type.

We used Cloud TPU v2 accelerators for both training and inference. In our experiments, each training epoch took on average around 4 minutes for BERT, 6 minutes for BERT Multiple Choice, and 15 to 25 minutes for T5 models.

\section{Development Set Results}
\label{sec:dev_set_results}
We reported the test set results in multiple settings in Section \ref{sec:experiments}. In this section, we report all those results on the development sets.

Table \ref{tab:main_res_dev} shows the development set results of different models for all domains and input types. We note that the general trends are very similar to that of the test sets. On average, the results of different models are slightly higher for the development set compared to the test set (up to $2.35\%$). This is expected as we have tuned the hyper-parameters on the development sets.

\section{Analyzing Correctly Classified Examples}
\label{sec:correct_analysis}
\changeB{We analyzed 100 correctly classified examples in the {\sc music} domain and assigned one or more categories (e.g., {\it date} or {\it genre}) to each example. We used the predictions of our T5 model with the {\sc Unshown Background} input.} \changeB{Table \ref{tab:correct_analysis} shows the results which cover a wide range of categories.}

\begin{table}
\centering
\small
\begin{tabular}{| c |} \hline
{\bf \textsc{books}} \\ 
``Remember that book we saw at the store?'' \\
``Hey, about that book I lent you last month...'' \\
``Can you get me that book on the first shelf?'' \\
``I really liked that book from the reading club...'' \\
``That book I got was super interesting!'' \\ \hline 
{\bf \textsc{music}} \\
``So that song I keep singing...'' \\
\makecell{``One of those cool songs that Bob\\sang last night...''}\\
``You sang that song really well yesterday...''\\
``Could you play that song from your playlist?'' \\
``I'll now play my favorite song.'' \\ \hline
{\bf \textsc{recipes}} \\
``Remember that fabulous stuff from Tom’s party?''\\
``That recipe on today’s Masterchef was too good!''\\
``Going to make that dish from Mary’s potluck.''\\
\makecell{``Our favorite food blogger had a cool \\ episode this week!''}\\
``Does mom's cookbook have that recipe?'' \\ \hline
\end{tabular}
\normalsize
\caption{The manual utterances which are used to populate the first cell of the cartoon.}
\label{tab:domain_questions}
\end{table}

\begin{table*}[t!]
\centering
\small
\begin{tabular}{|p{0.34\linewidth} | p{0.6\linewidth}|} \hline
{\bf Filter} & {\bf Rationale}\\ \hline
Articles with more than one infobox & Items should focus on a single topic. For example, we do not accept a movie that has a recorded song for the {\sc music} domain. \\ \hline
Items with a selected section length $\leq 250$ characters\footnote{For the {\sc music} domain, we always tested this constraint on the ``main'' Wikipedia section since we do not show Wikipedia sections as background knowledge for this domain.} & Items have enough information in the section selected to show as background knowledge to the annotators. \\ \hline
Books or music items that do not have genres in their infobox & Items contain important attributes for the domain \\ \hline
Recipes that are not a prepared food or without images (\S \ref{sec:item_background}) & Items contain  important attributes for the domain \\ \hline
Items in the {\sc music} domain with $\leq 14$ sections & Song should be popular to enable the annotators to also use their own background knowledge. \\ \hline
Items containing words on a denylist & Avoid sensitive or inappropriate items. \\ \hline
\end{tabular}
\normalsize
\caption{List of filters applied to select candidate items from those extracted from Wikipedia articles. For each filter, we show the rationale behind it.}
\label{tab:extraction_filters}
\end{table*}

\begin{table*}[t!]
\centering
\small
\begin{tabular}{| l | l | p{.65cm} | p{.65cm} | p{.65cm} | p{.65cm} || p{.65cm} | p{.65cm} | p{.65cm} |  p{.65cm} ||  p{.65cm} | p{.65cm} |  p{.65cm} |}
\multicolumn{2}{c}{\textsc{ }}& \multicolumn{4}{c ||}{\bf \textsc{Books}}  & \multicolumn{4}{c ||}{\bf \textsc{Recipes}} & \multicolumn{3}{c|}{\bf \textsc{Music}} \\ \hline
& & \textsc{Orac} & \textsc{Name} & \textsc{InBo} & \textsc{UnBa} & \textsc{Orac} & \textsc{Name} & \textsc{InBo} & \textsc{UnBa} & \textsc{Name} & \textsc{InBo} & \textsc{UnBa} \\ \hline
\multirow{3}{*}{\rotatebox[origin=c]{0}{\parbox{1cm}{\bf BERT}}} &  lr & 3e-5 & 1e-5 & 5e-6 & 1e-5 & 5e-6 & 5e-7 & 1e-5 & 3e-5 & 1e-5 & 3e-6  & 5e-6 \\ \cline{2-13}
& bsz & 16 & 16 & 32 & 16 & 16 & 16 & 32 & 64 & 64 & 64 & 32   \\ \cline{2-13}
& epochs & 5 & 10 & 3 & 3 & 3 & 1 & 3 & 1 & 1 & 3 & 3 \\ \hline
\multirow{3}{*}{\rotatebox[origin=c]{0}{\parbox{1.1cm}{\bf BERT \\ Multiple \\ Choice}}} &  lr & 3e-5 & 5e-6  & 3e-5 & 3e-5 & 3e-5 & 1e-6 & 3e-5 & 3e-5 & 5e-6 & 1e-5 & 5e-6 \\ \cline{2-13}
& bsz & 64 & 32 & 32 & 64 & 64 & 32 & 64 & 64 & 64 & 32 & 32  \\ \cline{2-13}
& epochs & 3 &  3 & 1 & 1 & 1 & 1 & 1 & 1 & 1 & 1 & 3 \\ \hline
\multirow{2}{*}{\rotatebox[origin=c]{0}{\parbox{1cm}{\bf T5}}} &  lr & 5e-6 & 3e-5 & 3e-6 & 3e-6 & 3e-6 & 3e-6 & 3e-6 & 3e-6 & 3e-6 & 3e-6  & 3e-6 \\ \cline{2-13}
& bsz & 64 &  32 & 64 & 64 & 32 &  32 & 16 & 64 & 64 & 64 & 32 \\ \hline
\end{tabular}
\normalsize
\caption{Selected hyper-parameters for the supervised models for each domain and input type. We list selected values for base learning rate (lr), Training batch size (bsz), Num training epochs (epochs).}
\label{tab:hyper-params}
\end{table*}

\begin{table*}[t!]
\centering
\small
\begin{tabular}{| l | p{.69cm} | p{.69cm} |  p{.69cm} | p{.69cm} || p{.69cm} | p{.69cm} | p{.69cm} |  p{.69cm} ||  p{.69cm} |  p{.69cm} | p{.69cm} || p{.69cm} |}
\multicolumn{1}{c}{\textsc{ }}& \multicolumn{4}{c ||}{\bf \textsc{Books}}  & \multicolumn{4}{c ||}{\bf \textsc{Recipes}} & \multicolumn{3}{c||}{\bf \textsc{Music}} & \multicolumn{1}{c}{\textsc{ } } \\ \hline
& \textsc{Orac} & \textsc{Name} & \textsc{InBo} & \textsc{UnBa} & \textsc{Orac} & \textsc{Name} & \textsc{InBo} & \textsc{UnBa} & \textsc{Name} & \textsc{InBo} & \textsc{UnBa} & \textsc{Avg} \\ \hline
USE & 66.06 & 55.15 & 59.12 & 58.41 & 70.77 & 52.48 & 64.98 & 66.36 & 57.53 & 60.71 & 60.57 & 61.10 \\ \hline
Entailment & 85.00 & 50.91 & 63.16 & 70.54 & 81.31 & {\bf 56.73} & 69.41 & 75.58 & 52.68 & 62.42 & 74.32 & 67.46 \\ \hline
BERT & 94.34 & 59.58 & 78.27 & 81.91 & 88.87 & 53.99 & 76.15 & 81.07 & {\bf 60.57} & 63.35 & 74.50 & 73.87 \\ \hline
BERT Joint & 95.00 & {\bf 61.85} & 77.31 & 82.47 & 89.58 & 56.60 & 76.86 & 81.21 & 59.79 & 68.07 & 76.17 & 74.99 \\  \hline
T5 & {\bf 95.91} & 61.04 & {\bf 78.98} & {\bf 84.13} & {\bf 93.22} & 56.69 & {\bf 82.80} & {\bf 85.77} & 59.14 & {\bf 72.33} & {\bf 82.97} & {\bf 77.54} \\
\hline
\end{tabular}
\normalsize
\caption{Indirect reference resolution development set results for different models on all domains and input types: {\sc Oracle} ({\sc Orac}), {\sc Name}, {\sc Infobox} ({
\sc InBo}), {\sc Unshown Background} ({\sc UnBa}). The best result of each column is boldfaced.}
\label{tab:main_res_dev}
\end{table*}

\begin{table*}[t!]
\centering
\small
\begin{tabular}{|p{0.16\linewidth} | p{0.32\linewidth}| p{0.31\linewidth}| p{0.09\linewidth}|} \hline
{\bf Category} & {\bf Example 1} & {\bf Example 2} & {\bf Percentage}\\ \hline
Date & was released in 2012 & the song that's only a few years old & 25\% \\\hline
Content & Singer compared his new life and the old. & Not the sad song & 24\% \\\hline
Singer or band & The one by a male singer & song is by an Irish rock band & 19\% \\\hline
Genre & It is the song that is R\&B. & it's that baroque pop ballad track & 13\% \\\hline
Further song info & Was remixed in the late 80s & The one sampled from Shirly Bassey & 10\% \\\hline
Comparison & The newer one & Released later & 10\% \\\hline
Negation & Not the song about greed & No not the one with Rap & 10\% \\\hline
Instrument or sound & It is a midtempo R\&B ballad & not the one with the piano intro & 7\% \\\hline
Album & One from their second album & The one from the album Wordshaker & 5\% \\\hline
\end{tabular}
\normalsize
\caption{Categories of correctly classified REs in the {\sc music} domain. The results are based on the T5 model with the {\sc Unshown Background} input.}
\label{tab:correct_analysis}
\end{table*}

\end{document}